\documentclass[sigconf]{acmart}
\AtBeginDocument{%
  }

\usepackage{booktabs}   
\usepackage{array}      
\usepackage{threeparttable} 
\usepackage{multirow}
\usepackage{multicol}
\usepackage{tabularx}
\usepackage{amsmath}
\usepackage{float}
\usepackage{placeins}
\usepackage{stfloats}

\copyrightyear{2026}
\acmYear{2026}
\setcopyright{cc}
\setcctype{by}
\acmConference[WWW Companion '26]{Companion Proceedings of the ACM Web Conference 2026}{April 13--17, 2026}{Dubai, United Arab Emirates}
\acmBooktitle{Companion Proceedings of the ACM Web Conference 2026 (WWW Companion '26), April 13--17, 2026, Dubai, United Arab Emirates}
\acmPrice{}
\acmDOI{10.1145/3774905.3793134}
\acmISBN{979-8-4007-2308-7/2026/04}
\begin{document}

\title{Click-to-Ask: An AI Live Streaming Assistant with Offline Copywriting and Online Interactive QA}

\author{Ruizhi Yu\textsuperscript{*}}
\email{51265902122@stu.ecnu.edu.cn}
\affiliation{%
\institution{East China normal University}
  \city{Shanghai}
  \country{China}
}
\author{Keyang Zhong\textsuperscript{*}}
\email{zhongky23@mail2.sysu.edu.cn}
\affiliation{%
\institution{Sun Yat-sen University}
  \city{Guangzhou}
  \country{China}
}
\author{Peng Liu\textsuperscript{$\dagger$}}
\email{arlen.liu@oppo.com}
\affiliation{%
\institution{OPPO AI Center}
  \city{Shenzhen}
  \country{China}
}
\author{Qi Wu}
\email{wuqi@oppo.com}
\affiliation{%
\institution{OPPO AI Center}
  \city{Shenzhen}
  \country{China}
}
\author{Haoran Zhang}
\email{hr.zhang3@siat.ac.cn}
\affiliation{%
\institution{Shenzhen Institutes of Advanced Technology}
  \city{Shenzhen}
  \country{China}
}
\author{Yanhao Zhang}
\email{zhangyanhao@oppo.com}
\affiliation{%
\institution{OPPO AI Center}
  \city{Shenzhen}
  \country{China}
}

\author{Chen Chen}
\email{chenchen4@oppo.com}
\affiliation{%
\institution{OPPO AI Center}
  \city{Shenzhen}
  \country{China}
}
\author{Haonan Lu}
\email{luhaonan@oppo.com}
\affiliation{%
\institution{OPPO AI Center}
  \city{Shenzhen}
  \country{China}
}
\renewcommand{\shortauthors}{Ruizhi Yu et al.}

\renewcommand{\shortauthors}{Yu et al.}

\begin{abstract}

Live streaming commerce has become a prominent form of broadcasting in the modern era. To facilitate more efficient and convenient product promotions for streamers, we present Click-to-Ask, an AI-driven assistant for live streaming commerce with complementary offline and online components. The offline module processes diverse multimodal product information, transforming complex inputs into structured product data and generating compliant promotional copywriting. During live broadcasts, the online module enables real-time responses to viewer inquiries by allowing streamers to click on questions and leveraging both the structured product information generated by the offline module and an event-level historical memory maintained in a streaming architecture. This system significantly reduces the time needed for promotional preparation, enhances content engagement, and enables prompt interaction with audience inquiries, ultimately improving the effectiveness of live streaming commerce. On our collected dataset of TikTok live stream frames, the proposed method achieves a Question Recognition Accuracy of 0.913 and a Response Quality score of 0.876, demonstrating considerable potential for practical application. The video demonstration can be viewed here: https://www.youtube.com/shorts/mWIXK-SWhiE.
\end{abstract}





\maketitle

\begin{figure*}[t]
\vspace{-3mm}
  \centering
  \includegraphics[width=0.9\textwidth]{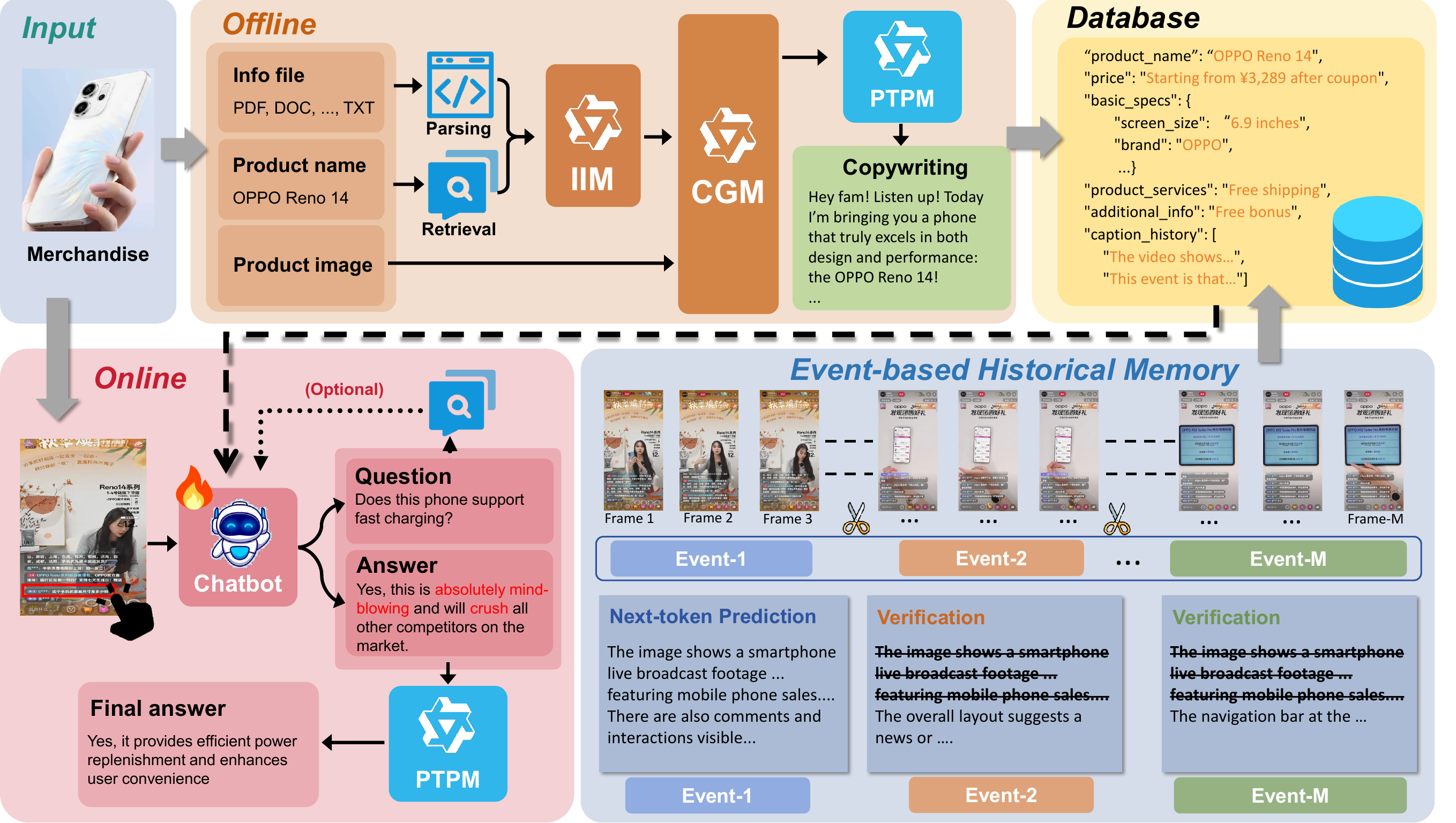}
  \vspace{-2mm} 
  \caption{Framework of Click-to-Ask. The upper half illustrates the offline copywriting module, including product information integration, copywriting generation, prohibited-term purification, and the structured product database. The lower half shows the online interactive Q\&A module on the left, responsible for real-time bullet-chat responses, and the event-based history memory module on the right, which asynchronously segments the video stream and accelerates historical caption generation.}
  \label{fig:arch}
\vspace{-3mm}  
\end{figure*}

\section{Introduction}
Live Streaming Commerce (LSC), which combines real-time video broadcasting with instant purchasing, has become a major force in digital retail. Streamers present and trial products while interacting with viewers, who can watch, ask questions, and purchase directly through links in the live interface. This model transforms traditional “person-to-product” shopping into an interactive “person-to-person” experience, blending entertainment, social interaction, and shopping. However, live selling often demands substantial preparation and careful management of product information.

The rise of AI offers new opportunities for automating content creation. Large language models (LLMs), with their text generation, knowledge integration, and Q\&A capabilities, can assist streamers in generating promotional content and handling viewer inquiries. Despite this potential, current AI-powered live-streaming assistants are limited. More efficient systems are needed to reduce streamers’ workload, enhance content appeal, and boost viewer engagement, accelerating AI adoption in LSC.

In this demo, we present the design and implementation of a Live Streaming Assistant System aimed at addressing two key bottlenecks in live commerce: the substantial time cost of crafting high-quality promotional copy and the difficulty of rapidly accessing product-specific information to respond to audience inquiries in real time. The system is organized into two tightly integrated components: an offline module and an online module.
In the offline module, the system first retrieves and extracts relevant product information from multiple sources. An Information Integration Module then fuses these retrieved materials with product documentation to construct structured product knowledge, which is subsequently stored in a database. This structured information is then fed into a Copywriting Generation Module, followed by a Prohibited-term Purification Module, to automatically generate engaging, regulation-compliant live streaming promotional copy.
In the online module, when a streamer clicks on a viewer’s question during a live session, the system captures the query in the context of the ongoing video stream and leverages both the pre-constructed structured product information and event-level historical memory to generate an appropriate response via a click-based chatbot. Before being delivered to the audience, the response is further screened by the prohibited-term purification module to ensure compliance with platform policies and live-streaming regulations.

\section{System Framework}
We illustrate the overview framework of our system in Figure 1. 
The system is comprised of two primary components: an offline copywriting module and an online interactive Q\&A module, which are dedicated to generating promotional copywriting and handling real-time bullet-chat responses, respectively.
Detailed descriptions of each will be provided in the following subsections.

\subsection{Offline copywriting module}
The offline module is designed to ingest heterogeneous product materials provided by streamers, transform them into structured product knowledge, and automatically generate persuasive promotional copy that adheres to live streaming regulations. The end-to-end pipeline comprises three core stages: product information integration, copywriting generation, and prohibited-term purification.

\subsubsection{Product information integration}
In this stage, user inputs are processed in multiple forms, including text, images, and files. File inputs support various formats such as PDF, DOC, XLSX, TXT, and WAV. 
Notably, PDF parsing is powered by our proprietary MinerU engine~\cite{wang2024mineru}, which integrates OCR and layout analysis to handle complex document structures, while WAV files are processed through Automatic Speech Recognition(ASR) to convert speech into text. 
Meanwhile, the product name provided by the streamer is used to retrieve external knowledge as supplementary information. 
Recognizing that user-provided data and external supplementary information often contain substantial redundancy and extraneous content irrelevant to the target product, their direct application in copywriting generation leads to suboptimal performance. To mitigate this issue, we introduce a Large Language Model-based Information Integration Module (IIM). This module is designed to perform three critical functions: extraction of salient elements, elimination of repetitive content, and filtration of irrelevant noise. The output of this process is a refined set of concise and structurally organized product descriptions. These refined descriptions are subsequently stored in a database to facilitate downstream copywriting generation tasks. 
Through Chain-of-Thought (CoT) prompt engineering, the model first extracts product-related information from the user-provided data, preserving core attributes. It then supplements this by retrieving and integrating relevant information pertaining to the same product from external supplementary sources. The final output is organized into a structured format comprising the following dimensions: product name, price, essential specifications, key features, and service details. 

\begin{table*}[!t]
\centering
\caption{Performance comparison of different methods on the test set.}
\label{tab:performance}
\vspace{-1mm} 
\begin{tabularx}{\linewidth}{l *{6}{>{\centering\arraybackslash}X} c}   
\toprule
\multirow{2}{*}{Baseline} &
\multicolumn{6}{c}{RQ} &
\multirow{2}{*}{QRA} \\
\cmidrule(lr){2-7}
 & electronics & appliances & clothing & food & others & mean & \\
\midrule
Qwen2.5-VL 7B         & 0.361 & 0.333 & 0.758 & 0.774 & 0.500 & 0.536 & 0.512 \\
+GRPO                 & 0.479 & 0.349 & 0.808 & 0.758 & 0.628 & 0.599 & 0.619 \\
+visual prompts       & 0.493 & 0.420 & 0.875 & 0.750 & 0.600 & 0.620 & 0.591 \\
+visual prompts +GRPO & 0.872 & 0.835 & 1.000 & 0.823 & 0.887 & 0.876 & 0.913 \\
\bottomrule
\vspace{-3mm} 
\end{tabularx}
\end{table*}

\subsubsection{Copywriting generation} 
Following the previous stage, the Copywriting Generation Module (CGM), which is powered by a Large Vision-Language Model (VLM), processes the structured product information and product images to generate copywriting. The module generates copy in various styles including literary, professional, and general tones. The prompts specifically guide the model to incorporate scenario-based narratives and emotional cues. This approach helps create warm, nostalgic narratives that resonate deeply with consumers. Additionally, the system provides commonly used live-stream interaction phrases to enhance streamer-viewer engagement and stimulate purchase desire. 

To deliver a more personalized experience, we have designed a self-adaptive copy generation pipeline that enables the model to emulate user-defined writing styles, thereby enhancing the overall user experience.

\subsubsection{Prohibited-term purification}
To ensure compliance with live streaming regulations, a Prohibited-term Purifier Module(PTPM) is designed to modify or remove non-compliant content based on a lexicon derived from TikTok’s community guidelines. This process effectively filters inappropriate expressions, enabling the generated copy to be used directly in broadcasts without further modification. 
Through prompts explicitly specifying definitions of common prohibited terms and guiding a step-by-step (CoT) reasoning process, the model first identifies any problematic phrases according to the provided definitions and then revises or removes them while maintaining semantic coherence and fluency. 

\subsection{Online interactive Q\&A module}
The online module assists streamers in quickly and accurately responding to viewer questions during live commerce broadcasts. When viewers post comments or bullet messages, streamers can click on the specific message to address. The system captures the corresponding video frame and click coordinates, which are processed by a click-based chatbot to extract the question. Responses are generated by integrating the question with structured product data from the database.To enhance answer accuracy for bullet-screen questions, the evaluation module can conditionally trigger a supplementary retrieval stage that leverages the question text itself to acquire additional context. All replies are then filtered through a prohibited-term purifier before delivery.



\begin{figure}[htbp]
\vspace{-3mm}
  \centering
  \includegraphics[width=\linewidth]{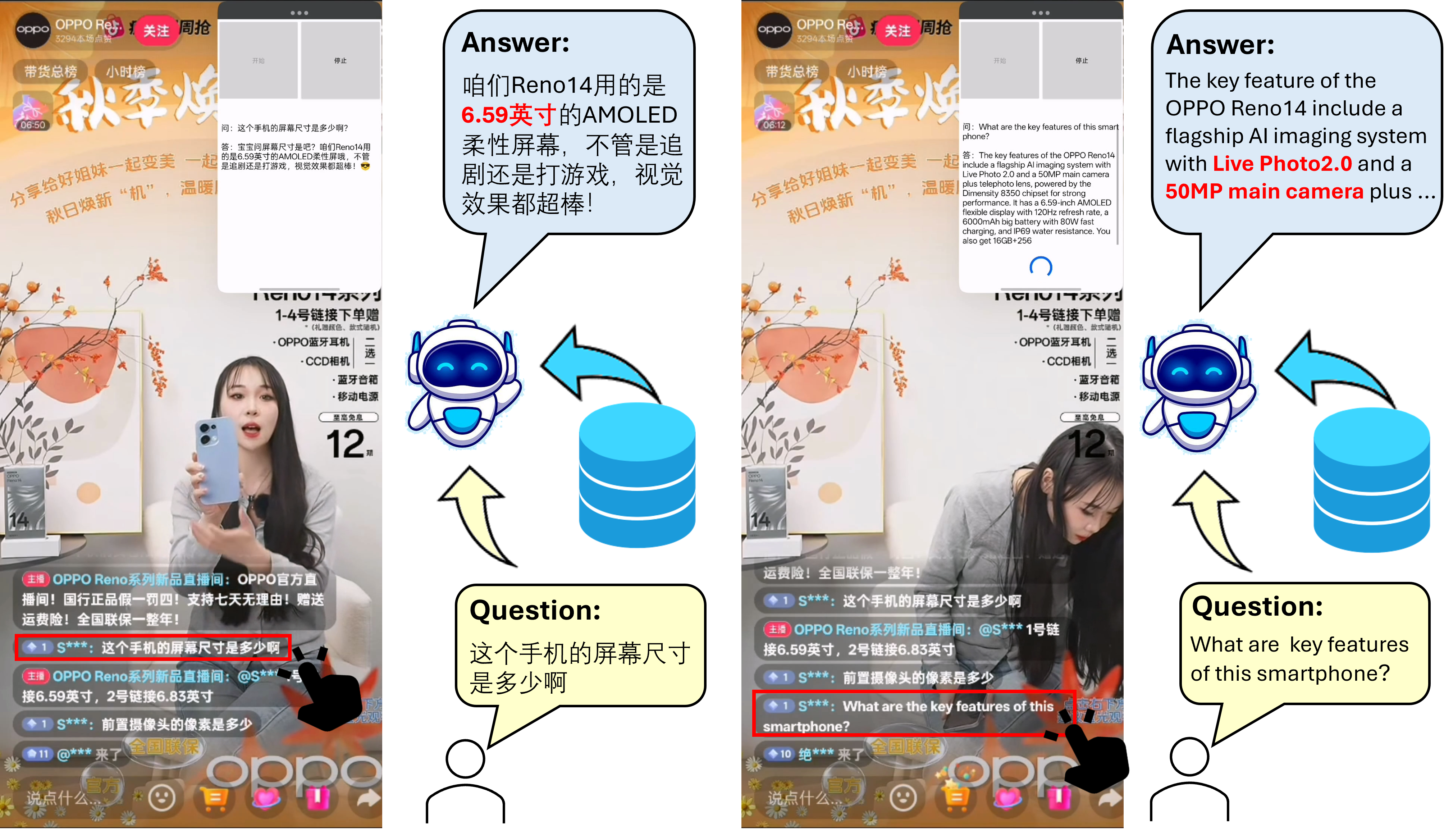}
  \vspace{-6mm}
  \caption{Some visual results on mobile devices.}
  \label{fig:arch}
\vspace{-4mm}  
\end{figure}

\subsubsection{Training Data Synthesis}
Empirical observations reveal that prevailing models struggle to localize the intended question from the click-position cue, leading to erroneous responses. To remedy this deficiency, we construct a targeted training set by repurposing the CLEVR dataset \cite{johnson2017clevr}, explicitly designed to enhance the model’s ability to identify and answer queries conditioned on precise click-position information.
In the original dataset, each image contains multiple QA pairs. We embed four question texts into each image to simulate viewer bullet messages in a live stream scenario. For each embedded text region, a random pixel coordinate is selected to simulate user click behavior. The resulting dataset consists of 8K images and 32K QA pairs. Each sample includes click coordinates as the question, and the corresponding on-screen text along with its correct response as the answer.

\subsubsection{Click-based streaming response learning}
We train the click-based chatbot using Group Relative Policy Optimization (GRPO)~\cite{shao2024deepseekmath} for 200 iterations on 8 H20 GPUs. As illustrated in Equation 1, a customized reward function is designed to incentivize accurate text recognition at the clicked location, with additional reward granted for generating appropriate responses based on the extracted text. The equation is as follows:
\begin{equation}
R(\hat{q}, \hat{a}, q^*, a^*)=
\begin{cases}
1, & \text{if } \hat{q} = q^* \land \hat{a} = a^* \\[4pt]
0.5, & \text{if } \hat{q} = q^* \land \hat{a} \neq a^* \\[4pt]
0, & \text{otherwise},
\end{cases}
\end{equation}
where $R$ denotes the reward function that assesses the quality of the model's output. Here, $\hat{q}$ and $\hat{a}$ represent the question and answer texts generated by the model, respectively, while $q^*$ and $a^*$ denote the corresponding ground-truth labels.
The correctness of the identified question text is evaluated via string matching, while the accuracy of the generated answer is assessed using GPT-4o.

Experimental results show that VLMs struggle to associate textual coordinate prompts with corresponding image locations. To address this, visual prompts are incorporated during training by superimposing a mouse cursor icon at each click point, enhancing multimodal alignment between coordinates and visual context. This approach is compared with other designs, and the results are presented in the section \ref{sec:experiments}. The result demonstrates that incorporating visual prompts at the click coordinates significantly enhances the model's ability to comprehend the relationship between coordinate inputs and fine-grained image content.

\subsubsection{Event-based accelerated segment for history memory storage}
To tackle information redundancy and model overload in lengthy live streams, we propose a plug-and-play Streaming Event Segmentation (SES) module which continuously divides the video stream into several logically complete event units based on semantic coherence and inherent causal relationships by processing input frames in a sliding window manner. Within each window, the model assesses the semantic relevance between adjacent frames based on the ViT feature similarity $c_t^{\text{ViT}}$ and the average optical flow magnitude $m_t$.  The similarity feature for each frame is computed as:
\begin{equation}
\hat{c_t} = \gamma c_t^{\text{ViT}} + (1 - \gamma) m_t, \quad \gamma=0.5
\end{equation}
Based on the feature similarity, the depth estimate for the $i$-th frame is calculated as:$d_i = \frac{\hat{c_{l_i}} + \hat{c_{r_i}} - 2\hat{c_i}}{2}$
where $\hat{c_{l_i}}$ and $\hat{c_{r_i}}$ represent the left and right peak values corresponding to the $i$-th frame, respectively. The model maintains the mean $\mu$ and variance $\sigma^2$ of $d_i$ within each sliding window. A segmentation boundary is identified when $
d_i > \mu + \alpha \cdot \sigma$ with $\alpha=1.0$.
Additionally, a plug-and-play Knowledge Extraction Accelerator (KEA) module is implemented to accelerate new fragment caption generation by calculating next-token prediction logits to select the optimal historical caption tokens as prefixes.
Let $L_c$ denote the length of the historical segment caption. This module selects an index from the sequence $[1, L_c-1]$ to indicate the specific position for caption continuation. For a given index $z \in [1, L_c-1]$, the module computes the average log probability $\mu_{[z-\delta:z-1]}$ of the preceding $\delta$ tokens and the log probability $l_z$ at the current index position. 
The truncation position is determined by the discrepancy between the average confidence in the historical window and the confidence score of the current token, defined as:
\begin{equation}
o = \min \left\{ z \in [\delta+1, L_c-1] \ \middle| \ 
\begin{aligned}
&\mu_{[z-\delta:z-1]} - l_z > \\
&\max(\alpha \cdot \mu_{[z-\delta:z-1]}, \beta)
\end{aligned} \right\}, \quad \beta = 0.8
\end{equation}
Here, $\max(\alpha \cdot \mu_{[z-\delta:z-1]}, \beta)$ acts as a dynamic threshold that quantifies the confidence discrepancy. The first position $o$ that exceeds this threshold is designated as the truncation point. This mechanism significantly reduces the information extraction burden for the model when generating event captions. Importantly, both SES and KEA modules are executed asynchronously in the background and have minimal impact on the speed and responsiveness of the main online module.

\section{Experiments}\label{sec:experiments}
We adopt Qwen3 8B~\cite{yang2025qwen3} as the LLM backbone for both the IIM and the PTPM, and use Qwen2.5-VL 7B~\cite{bai2025qwen2} as the base model for the click-based chatbot. We evaluate our system on a real-world TikTok-style dataset containing 396 QA pairs spanning five major product categories: electronics, appliances, clothing, food, and others.For the offline module, we assess the effectiveness of the PTPM. After integrating this module, the average number of prohibited terms in generated copywriting drops from 4.36 to 0.36, indicating a substantial improvement in content compliance.For the online module, we evaluate Question Recognition Accuracy (QRA) and Response Quality (RQ), where RQ is scored by GPT-4o against several methods (Table~\ref{tab:performance}). The Qwen2.5-VL 7B baseline achieves RQ = 0.536 and QRA = 0.512 but struggles to associate click coordinates with the corresponding bullet-chat messages. We first apply GRPO post-training with synthetic data using only textual coordinate prompts, which fails to sufficiently capture this spatial mapping. We then introduce visual prompts by overlaying a mouse cursor icon at the clicked location. Adding visual cues without further training yields moderate gains, while combining visual prompts with GRPO post-training (using both textual coordinates and visual markers) leads to the best performance, achieving RQ = 0.876 and QRA = 0.913. Our demo showcases several visual results on mobile devices, as shown in Figure 2.

\section{Conclusion}
In this demo, we presented Click-to-Ask — an AI Live Streaming Assistant with Offline Copywriting and Online Interactive Q\&A for live streaming commerce. The system addresses two key challenges for streamers: reducing the time cost of preparing product-promotion copywriting via an offline module, and enabling fast, accurate responses to viewer inquiries through an online click-based chatbot that leverages both pre-processed product information and event-level historical memory for context-aware bullet-chat Q\&A. To enhance model performance, we built a multimodal dataset that closely simulates real live streaming scenarios and conducted GRPO post-training, showing that integrating visual prompts with GRPO yields the best results. Deployed in real-world settings, Click-to-Ask delivers strong performance and a satisfactory user experience, demonstrating its practical value for live streaming commerce.
\bibliographystyle{ACM-Reference-Format}
\bibliography{ref}

\end{document}